\documentclass[10pt,twocolumn,letterpaper]{article}

\usepackage{cvpr}              
\definecolor{cvprblue}{rgb}{0.21,0.49,0.74}
\usepackage[pagebackref,breaklinks,colorlinks,allcolors=cvprblue]{hyperref}

\def\paperID{4} 
\def\confName{SAUAFG}
\def\confYear{2026}

\title{A Study of Failure Modes in Two-Stage Human–Object Interaction Detection}

\author{
{Lemeng Wang}$^{1}$\thanks{Equal Contribution. https://florawlm.github.io/DiagHOI/} \hspace{0.6em} 
{Qinqian Lei}$^{2*}$ \hspace{0.6em}
{Vidhi Bakshi}$^{1}$ \hspace{0.6em}
{Daniel Yi}$^{1}$ \hspace{0.6em}
{Yifan Liu}$^{1}$ \hspace{0.6em}
{Jiacheng Hou}$^{1}$ \\
{Asher Seng Hao}$^{4}$ \hspace{0.6em}
{Zheda Mai}$^{1}$ \hspace{0.6em}
{Wei-Lun Chao}$^{3}$ \hspace{0.6em}
{Robby T. Tan}$^{2}$ \hspace{0.6em}
{Bo Wang}$^{5}$ \\
$^{1}$The Ohio State University \quad
$^{2}$National University of Singapore \quad
$^{3}$Boston University \\
$^{4}$Independent Researcher \quad
$^{5}$University of Mississippi \\
{\tt\small wang.15543@buckeyemail.osu.edu \quad qinqian.lei@u.nus.edu \quad hawk.rsrch@gmail.com}
}

\begin{document}
\maketitle

\begin{abstract}
Human–object interaction (HOI) detection aims to detect interactions between humans and objects in images. While recent advances have improved performance on existing benchmarks, their evaluations mainly focus on overall prediction accuracy and provide limited insight into the underlying causes of model failures. 
In particular, modern models often struggle in complex scenes involving multiple people and rare interaction combinations. 
In this work, we present a study to better understand the failure modes of two-stage HOI models, which form the basis of many current HOI detection approaches. 
Rather than constructing a large-scale benchmark, we instead decompose HOI detection into multiple interpretable perspectives and analyze model behavior across these dimensions to study different types of failure patterns.
We curate a subset of images from an existing HOI dataset organized by human–object–interaction configurations (e.g., multi-person interactions and object sharing), and analyze model behavior under these configurations to examine different failure modes.
This design allows us to analyze how these HOI models behave under different scene compositions and why their predictions fail.
Importantly, high overall benchmark performance does not necessarily reflect robust visual reasoning about human–object relationships. 
We hope that this study can provide useful insights into the limitations of HOI models and offer observations for future research in this area.
\end{abstract}

\begin{figure}[t]
    \centering
    \includegraphics[width=\linewidth]{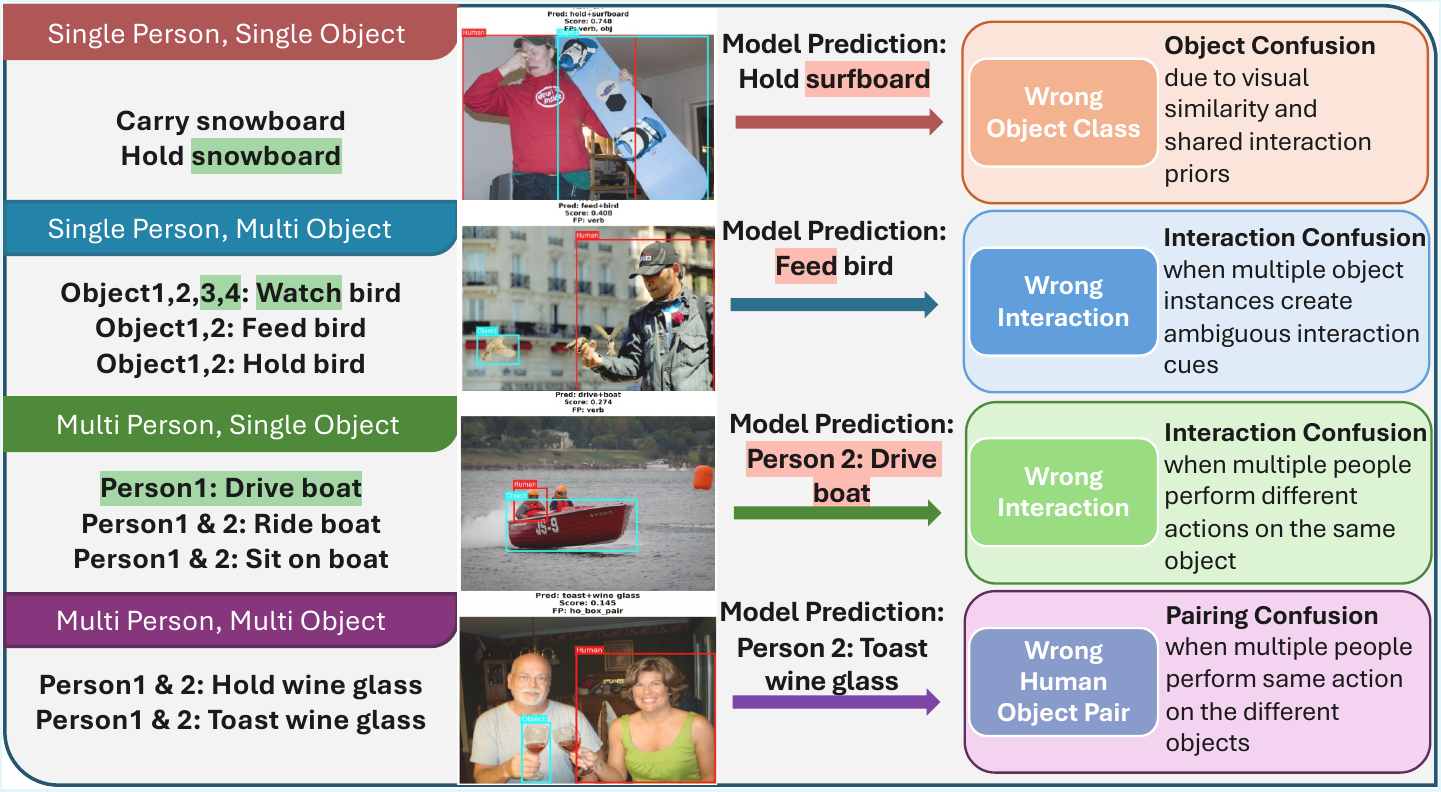}
    \caption{
    Qualitative examples of representative failure modes in HOI detection. 
    }
    \label{fig:teaser}
\end{figure}

\section{Introduction}
\label{sec:intro}

Human--object interaction (HOI) detection aims to detect how humans interact with objects in images, such as \emph{ride bicycle}, \emph{hold cup}, or \emph{read book}. As an important problem in human-centric visual understanding, HOI detection has relevance to applications such as activity understanding, assistive systems, robotics, and embodied AI~\cite{goodrich2008human,yao2012recognizing}. 
Over the past several years, the field has made promising progress, driven by advances in HOI benchmarks such as HICO-DET~\cite{hico}, V-COCO~\cite{gupta2015visual}, and SWiG-HOI~\cite{swighoi}, as well as increasingly powerful architectures for HOI detection~\cite{radford2021learning,li2022blip,li2023blip2,liu2024LLaVA,bai2025qwen25VL}.
%

Despite this progress, existing HOI evaluation protocols remain largely centered on aggregate detection metrics, such as mean Average Precision (mAP), which provide limited insight into why models succeed or fail. 
Recent studies have shown that HOI models, 
often struggle in complex scenarios involving multiple people, multiple objects, or subtle interaction distinctions, where interactions must be correctly attributed to the right human–object pair. For example, prior work has observed that multi-person scenes and cases where different subjects perform similar or distinct interactions are particularly challenging for HOI detection models~\cite{lei2026crosshoi_bench}.

However, existing benchmarks \cite{hico,gupta2015visual,swighoi,lei2026crosshoi_bench} typically do not explicitly analyze model behavior under these different interaction configurations.
In practice, current benchmarks do not explicitly separate different human–object interaction configurations, such as single-person versus multi-person scenes, interactions involving the same or different object instances, and cases where multiple subjects perform similar or distinct interactions.
As a result, errors arising from ambiguous human–object pairing and interaction attribution can be mixed together in evaluation results, making it difficult to understand the underlying causes of model failures. Consequently, strong overall benchmark performance does not necessarily imply robust visual reasoning about human–object relationships.

In this work, we study the behavior of two-stage HOI models from a structured analysis perspective. 
We focus on two-stage methods, where detection and interaction modeling are handled in separate stages, allowing clearer analysis of interaction-related failure modes.
Instead of proposing another large-scale benchmark, we analyze a curated subset of data to examine different human–object interaction configurations. 
Our goal is to analyze how two-stage HOI models behave under controlled variations in human–object interaction configurations, including changes in the number of people and objects, object sharing, and interaction consistency.
This analysis allows us to identify challenging interaction configurations and examine sources of prediction errors such as localization, interaction classification, and human–object pairing, which are often not visible from aggregate performance metrics.

To support this analysis, we curate a subset of images from an established HOI benchmark such as HICO-DET \cite{hico}, and organize them based on different aspects of scene composition.
This subset is not designed for scale, but for targeted analysis of model behavior under controlled yet practically meaningful conditions.
Using this setup, we analyze model behavior from two complementary perspectives. 
First, we study performance across interaction configurations defined by variations in the number of people, object sharing, and interaction consistency, as illustrated in \cref{fig:teaser}, which shows representative examples of these configurations and highlights typical error types, such as object confusion and human--object pairing errors.
Second, we analyze prediction errors across these configurations by decomposing model failures into interpretable components, including human detection, object detection, interaction prediction, and human–object pairing.
Together, these analyses provide a more interpretable view of current model limitations.
Our study makes the following contributions:
\begin{itemize}
    \item We present a structured analysis of two-stage HOI detection models, moving beyond aggregate metrics to study model behavior under different human--object interaction configurations.
    \item We analyze prediction errors across these configurations by decomposing model failures into interpretable components, including human detection, object detection, human--object pairing, and interaction prediction.
    \item We provide observations on HOI failure modes, including verb-related errors, failures in multi-person scenes with different interactions, and object-conditioned biases.
\end{itemize}

\section{Related Work}
\label{sec:related_work}
\noindent \textbf{HOI Detection Methods}
Existing approaches to HOI detection can be broadly grouped into two paradigms. 
Two-stage methods follow a detection-then-classification pipeline, where human and object instances are first localized and interaction labels are subsequently inferred for candidate pairs~\cite{zhang2021spatially, zhang2022efficient, park2023viplo,hou2020visual, lei2024EZ_HOI,hou2021detecting,lei2023efficient,mao2024clip4hoi,liu2022interactiveness,zhang2023exploring,lei2024exploring}. 
In contrast, one-stage methods aim to directly predict human–object–interaction triplets in a unified framework, typically leveraging transformer-based architectures for joint reasoning~\cite{zou2021end,qu2022distillation,li2024neural,wu2023end,ning2023hoiclip,hong2025learning,Liao_2022_CVPR,kim2023relational}.  
While these designs differ in modeling strategy, their performance is commonly evaluated using aggregate metrics such as mAP, which summarize overall performances but provide limited visibility into how errors arise. 
Consequently, it can be unclear whether improvements in benchmark scores reflect advances in interaction reasoning or are influenced by biases in data distribution or easier scenarios.

\noindent \textbf{Existing HOI Benchmarks and Evaluation}
Several benchmarks have been proposed for HOI detection, including HICO-DET~\cite{chao2018learning}, V-COCO~\cite{lin2014microsoft}, and SWiG-HOI~\cite{wang2022learning}, which differ in scale and label space.
Despite these variations, they largely rely on mean Average Precision (mAP) with exact matching between predicted and annotated HOI triplets.
While this protocol enables standardized comparison across methods, it mainly reflects aggregate performance and offers limited insight into the underlying causes of model errors.
In particular, it may not explicitly separate different sources of difficulty, such as ambiguity in human--object pairing, or errors in interaction recognition, especially in complex multi-person scenes.
Recent work such as CrossHOI-Bench~\cite{lei2026crosshoi_bench} and semantic evaluation~\cite{noack2026shoe} further explore HOI evaluation from different perspectives, including comparing vision-language models with HOI-specific approaches and incorporating semantic similarity for open-vocabulary evaluation.
However, these approaches primarily focus on overall performance comparison, without explicitly modeling scene structure or distinguishing types of incorrect predictions.
As a result, existing benchmarks provide limited visibility into how and why HOI models fail, motivating more fine-grained analysis of model behavior.

\begin{figure*}[t]
    \centering
    \includegraphics[width=\linewidth]{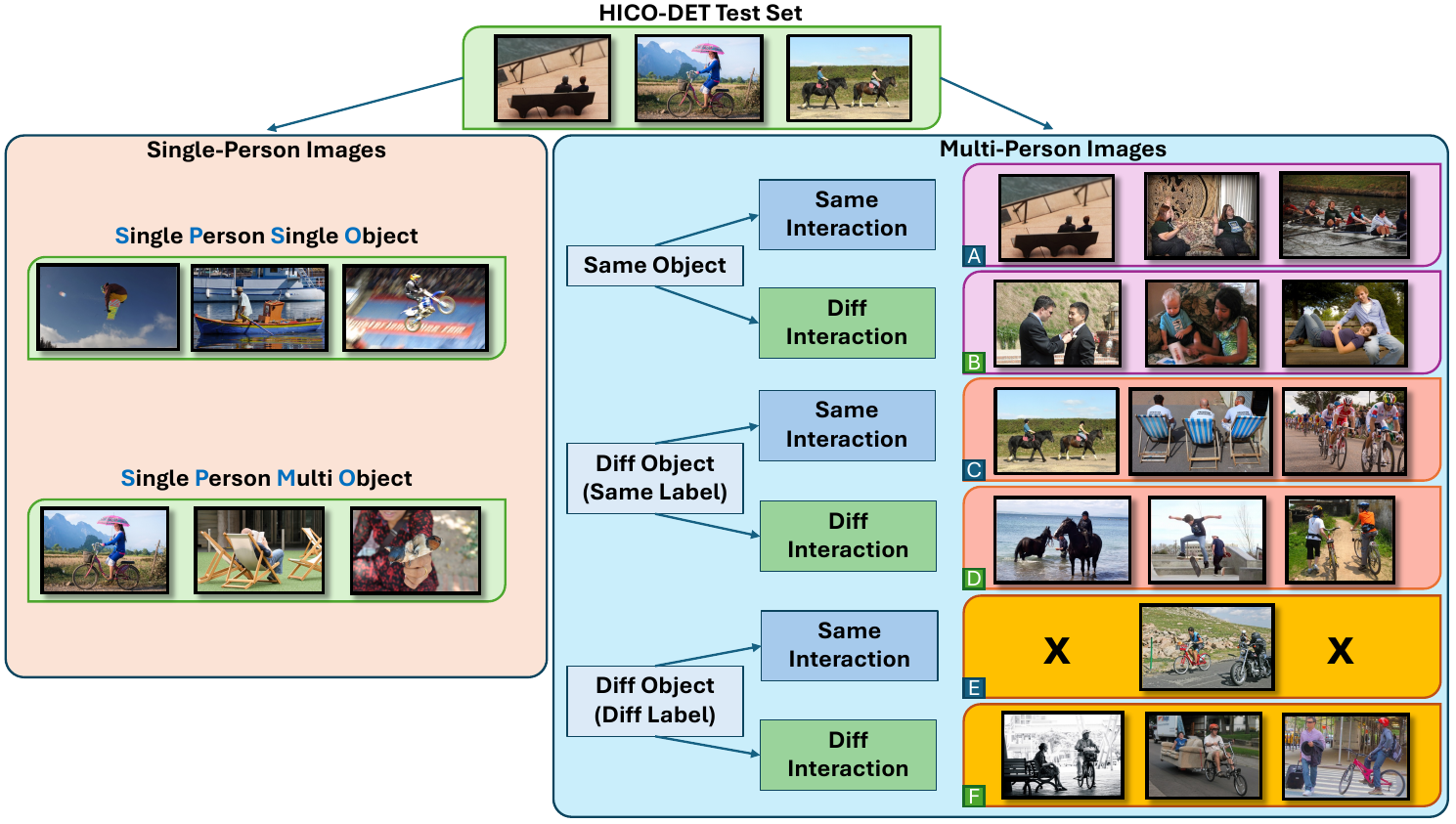}
    \caption{
    Overview of how we organize the HICO-DET test set for analysis. 
    We first divide images into \textbf{single-person} and \textbf{multi-person} scenarios. 
    For single-person images, we define two subsets: \textbf{single-person single-object (SPSO)} and \textbf{single-person multiple-object (SPMO)}, which isolate object ambiguity without subject ambiguity. 
    For multi-person images, we categorize samples along two axes: \textbf{object relation} (same vs. different object instances) and \textbf{interaction relation} (same vs. different interaction). Their combination yields six categories (A--F), capturing different levels of ambiguity in human--object pairing and interaction attribution.
    }
    \label{fig:taxonomy}
\end{figure*}

\section{Subset Construction}
\label{sec:method}

\subsection{Overview}
We construct a subset from the HICO-DET test set to analyze the failure modes of HOI detection models.
Instead of relying solely on aggregate evaluation metrics such as mean Average Precision (mAP), our goal is to analyze how two-stage HOI models behave under controlled variations in human--object interaction configurations.
In particular, we focus on configurations involving multi-person interactions, human--object pairing, and interaction diversity, which have been recognized as challenging scenarios in prior work \cite{lei2026crosshoi_bench}.
These factors frequently occur in real-world images but are not explicitly separated in existing benchmarks such as HICO-DET~\cite{hico}, V-COCO~\cite{gupta2015visual}, and SWiG-HOI~\cite{swighoi}, making it difficult to understand model behavior across different interaction configurations. As a result, it is even more challenging to identify the types of prediction errors that occur in each configuration.
To study these aspects, we reorganize the HICO-DET test set into structured subsets that capture different types of ambiguity in human--object interaction configurations.
This setup enables controlled analysis of model behavior under increasingly complex conditions, allowing us to move beyond overall performance and obtain a more interpretable understanding of model errors.

\noindent \textbf{Dataset Filtering and Scope}
We construct our benchmark from the full HICO-DET test set, with categories defined strictly based on its annotated human--object pairs. 
Specifically, we consider only human--object pairs that are explicitly annotated in HICO-DET when determining the category of each image. 
This restricts the label space to the annotated interactions and avoids ambiguity from unlabeled entities, ensuring a well-defined and consistent categorization.
To focus on visually meaningful interaction reasoning, we remove two types of images from the HICO-DET test set based on their annotations: (1) images whose annotated HOIs consist only of \texttt{no\_interaction}, and (2) images whose HOIs are all marked as invisible. These cases do not provide reliable visual evidence for evaluating human--object pairing or interaction attribution, as they lack observable interaction cues.
After filtering, the remaining images form the basis of our analysis.

\subsection{Hierarchical Organization}
\label{sec:taxonomy}
After filtering, we organize the dataset using a hierarchical structure based on the number of human subjects in each image. 
This organization reflects a key difference in interaction complexity: single-person scenes involve limited actions, while multi-person scenes are more complex due to the presence of multiple human--object pairs.
We first divide images into \textit{single-person} and \textit{multi-person} scenarios. Single-person images remove subject ambiguity and allow us to analyze object-related ambiguity in isolation, while multi-person images introduce ambiguity in human--object pairing and interaction attribution.

\noindent \textbf{Single-Person Cases}
For single-person images, we further divide them into two subsets:
\begin{itemize}
    \item \textbf{Single-person single-object (SPSO):} images containing one annotated person and one relevant annotated object.
    \item \textbf{Single-person multiple-object (SPMO):} images containing one annotated person and multiple relevant annotated objects.
\end{itemize}
The SPSO subset represents the simplest setting, where both subject and object ambiguity are minimal. In contrast, the SPMO subset introduces ambiguity in object selection, as multiple candidate objects are present for a single person. This hierarchical structure captures a progression of interaction complexity, where ambiguity increases from SPSO to SPMO and further to multi-person scenarios, enabling controlled analysis of different failure modes.

\noindent \textbf{Multi-Person Cases}
We now introduce our organization for multi-person scenarios, which constitute more complex cases for HOI detection. In such scenarios, models must correctly associate multiple humans with their corresponding objects and interactions, often involving ambiguity in human–object pairing and interaction attribution.
We define this organization based on two key factors: (1) the relationship between the objects associated with different people, and (2) whether the corresponding interactions are the same or different. The first factor captures whether subjects share the same object instance or interact with different objects, while the second factor captures variation in interaction semantics.
By combining these two axes, we obtain six categories (A--F), as illustrated in ~\cref{fig:taxonomy}:

\begin{itemize}
    \item \textbf{A:} the same object instance, same interaction
    \item \textbf{B:} the same object instance, different interaction
    \item \textbf{C:} different object instances with the same object label, same interaction
    \item \textbf{D:} different object instances with the same object label, different interaction
    \item \textbf{E:} different object instances with different object labels, same interaction
    \item \textbf{F:} different object instances with different object labels, different interaction
\end{itemize} 
Categories A and B correspond to scenarios in which multiple people interact with the same object instance. While category A represents relatively less ambiguous cases where all subjects perform the same interaction, category B introduces ambiguity in interaction attribution for a shared object.
Categories C and D correspond to cases where subjects interact with different object instances that share the same object label. These scenarios introduce ambiguity in distinguishing between visually similar objects, often leading to incorrect human--object pairing.
Categories E and F represent the most complex cases, where both object identity and interaction differ across subjects. These scenarios require models to jointly consider object semantics, spatial relationships, and interaction types. However, we observe that such cases are rare in HICO-DET, with only a small number of images falling into categories E and F. This reflects the natural distribution of scenes in the dataset rather than a limitation of our construction. As a result, our quantitative analysis primarily focuses on categories A--D, where sufficient data is available for reliable evaluation, while still including E and F for completeness.

\noindent \textbf{Definition of Same and Different Interactions.}
A key component of our taxonomy is distinguishing whether multiple people perform the same or different interactions. We define two subjects as having the \textit{same interaction} if they are associated with the same HOI category for their corresponding objects. We define interactions as \textit{different} in two cases: (1) when the subjects perform distinct HOIs (e.g., one person is \textit{holding} an object while another is \textit{throwing} it), or (2) when one subject performs an additional clearly identifiable action that is absent in the other. For example, if two people are both \textit{riding} a horse but only one is also \textit{holding} it, we categorize them as having different interactions. This definition is important because subtle asymmetries in interaction can significantly affect human--object pairing and attribution. Without explicitly capturing these differences, models may appear correct under coarse evaluation metrics while still making incorrect human–object associations. 
By adopting this definition, our organization captures fine-grained distinctions that are important for diagnosing model behavior. 
Overall, this setup allows us to analyze ambiguity along both object and interaction dimensions, enabling controlled analysis of model failure modes.

\subsection{Annotation Protocol}
To ensure the reliability of our category labels, we adopt a multi-annotator protocol. Each image is assigned a single diagnostic category label corresponding to our organization (i.e., SPSO, SPMO, and categories A–F shown in ~\cref{fig:taxonomy}), which is distinct from the original HOI annotations in HICO-DET.
Each image is independently labeled by three annotators according to the organization defined in ~\cref{fig:taxonomy}. 
Annotators are instructed to assign a category based only on ground-truth annotated persons and objects, following the definitions of object relation and interaction relation. The final category for each image is determined by majority vote. 
This process helps reduce subjectivity and improves consistency, especially in cases where distinguishing between same and different interactions may be subtle. 
In particular, annotators follow a predefined definition of interaction difference: interactions are considered different not only when the actions are distinct, but also when one subject performs an additional clearly identifiable action. This guideline ensures consistent handling of asymmetric interaction cases. 
For our analysis, we focus on images that receive a \textbf{single consensus label} after annotation. This filtering improves label reliability and enables controlled evaluation across categories.

\begin{figure}[t]
    \centering

    \begin{minipage}[t]{0.48\textwidth}
        \centering
        \includegraphics[width=\linewidth]{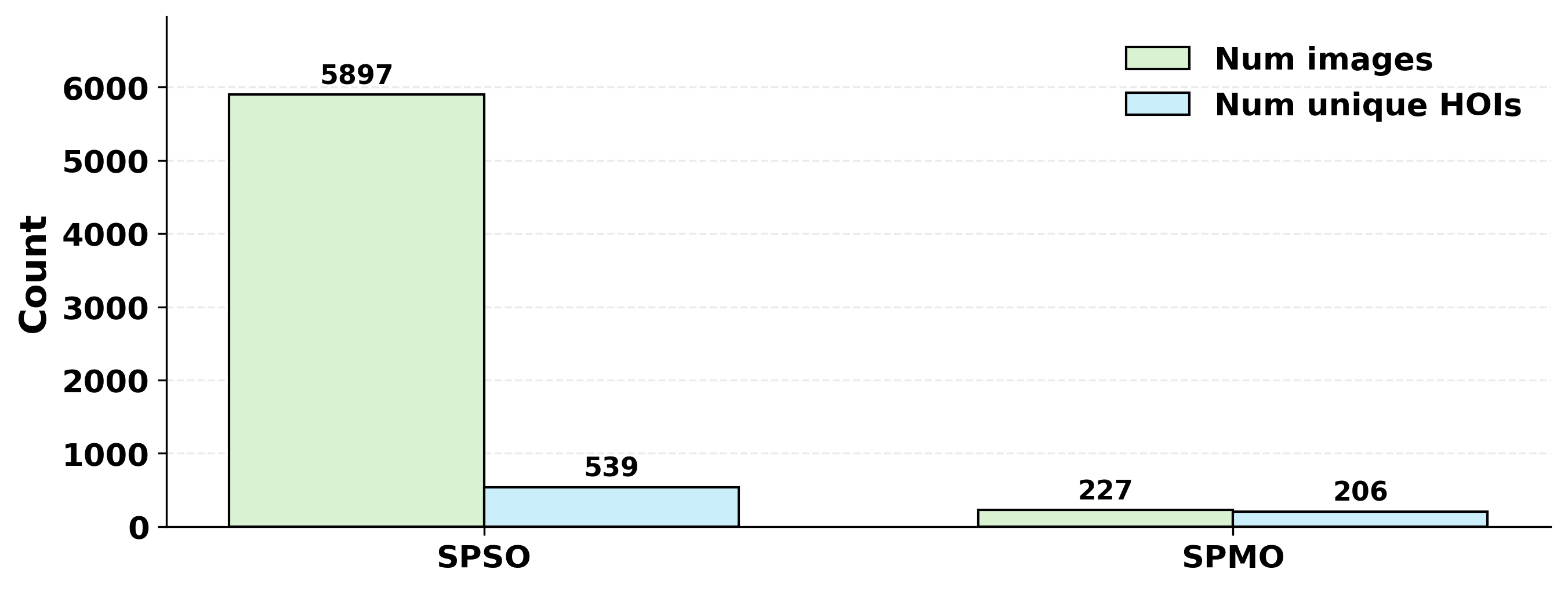}
        \\
        {\small (a) Single-person category distribution (SPSO vs. SPMO)}
    \end{minipage}
    \hfill
    \begin{minipage}[t]{0.48\textwidth}
        \centering
        \includegraphics[width=\linewidth]{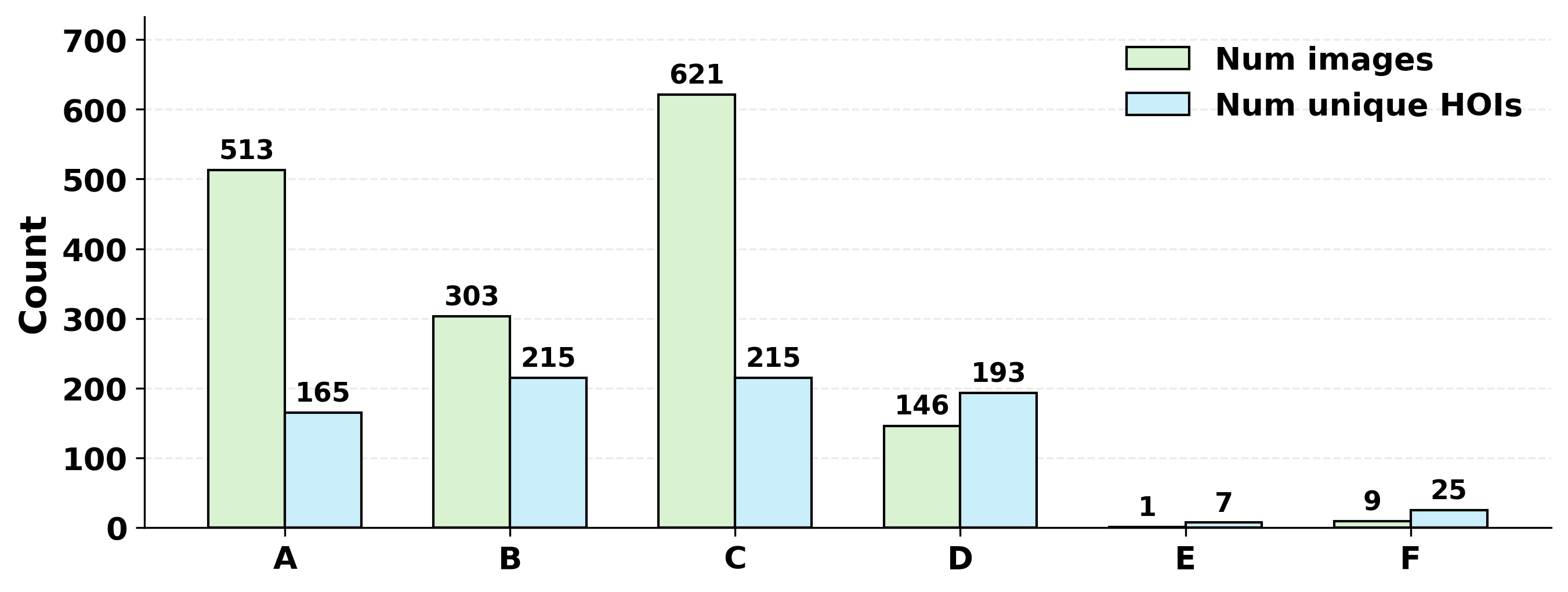}
        \\
        {\small (b) Multi-person category distribution (A--F)}
    \end{minipage}

    \caption{
   Distribution of images and HOIs across the subsets.
    }
    \label{fig:distribution}
\end{figure}

\subsection{Subset Statistics}
\label{Benchmark_stats}
We report the distribution of images and HOIs across the constructed subsets in ~\cref{fig:distribution}. 
For single-person images, the dataset is dominated by the SPSO subset (5,897 images), while the SPMO subset contains significantly fewer samples (227 images). In total, 6,124 out of 9,658 test images in HICO-DET contain only a single person. 
This suggests that standard evaluation metrics such as mAP are largely influenced by performance on single-person scenarios, where ambiguity in human--object pairing is minimal. For multi-person images, the distribution across categories A--F is highly imbalanced. Categories A (513 images) and C (621 images) contain the largest number of samples, followed by B (303) and D (146). In contrast, categories E (1 image) and F (9 images) are rare.

This imbalance reflects the natural distribution of scenarios in HICO-DET rather than a limitation of our subset construction. 
Different categories highlight different sources of error. 
Categories A and B capture shared-object scenarios, while categories C and D introduce ambiguity across multiple instances of the same object class, making correct human--object pairing more challenging and often leading to pairing errors. 
Given the limited number of samples in categories E and F, we do not draw quantitative conclusions from these subsets. Instead, we focus our analysis on categories with sufficient data (primarily A--D), which allow more controlled analysis of model behavior.

Overall, these observations suggest two characteristics of existing HOI benchmarks: (1) aggregate metrics are dominated by relatively simple single-person scenarios, and (2) more challenging multi-person cases are underrepresented but critical for understanding model failure modes. 
Our subset follows this structure and supports analysis of these challenging settings.

\section{Experiment}
\label{sec:exp}
\noindent \textbf{Baselines}
We evaluate four representative two-stage HOI detection models: ADA-CM~\cite{lei2023efficient}, CMMP~\cite{lei2024exploring}, HOLa~\cite{lei2025hola}, and LAIN~\cite{kim2025locality}. All models are evaluated in the fully-supervised setting using their official implementations and released best-performing checkpoints, with ADA-CM, CMMP, and HOLa based on ViT-L backbones and LAIN based on a ViT-B backbone. 
All models are evaluated using the same setup on the constructed subsets.

\noindent \textbf{Implementation Details}
We use the official released checkpoints for all evaluated models without additional training or fine-tuning. 
All models are evaluated on the HICO-DET test set, and their predictions are analyzed on the constructed subsets described in~\cref{sec:method}. 
For each image, the diagnostic category is determined based on ground-truth annotations, and model predictions are analyzed within the corresponding category. 
To ensure consistent analysis, all models are evaluated using the same evaluation process and matching rules. 
We restrict the benchmark to images with a single consensus category to avoid ambiguity across categories and ensure reliable category-wise analysis.

\begin{figure}[t]
    \centering
    \includegraphics[width=0.8\linewidth]{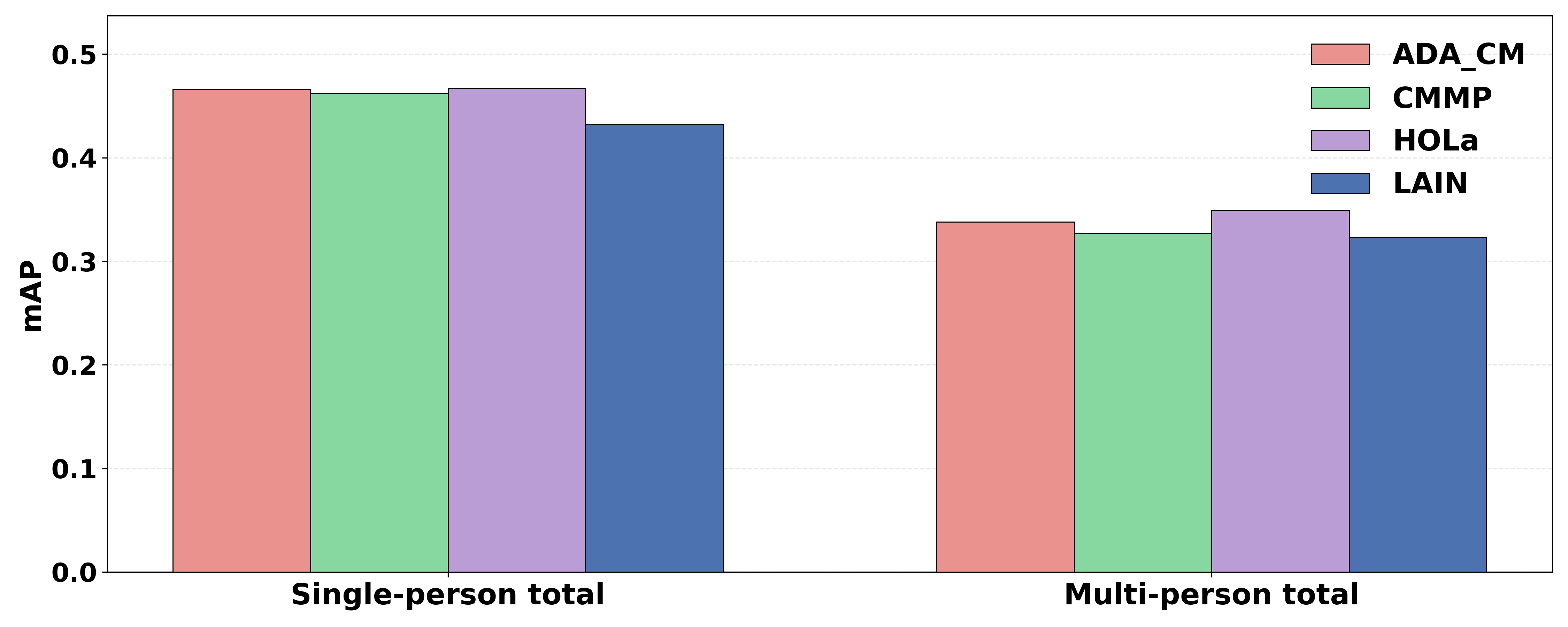}
    \caption{
    mAP comparison between single-person and multi-person scenarios. All models show a performance drop in multi-person settings, suggesting increased difficulty in this setting.
    }
    \label{fig:single_vs_multi}
\end{figure}

\begin{figure}[t]
    \centering
    \includegraphics[width=0.85\linewidth]{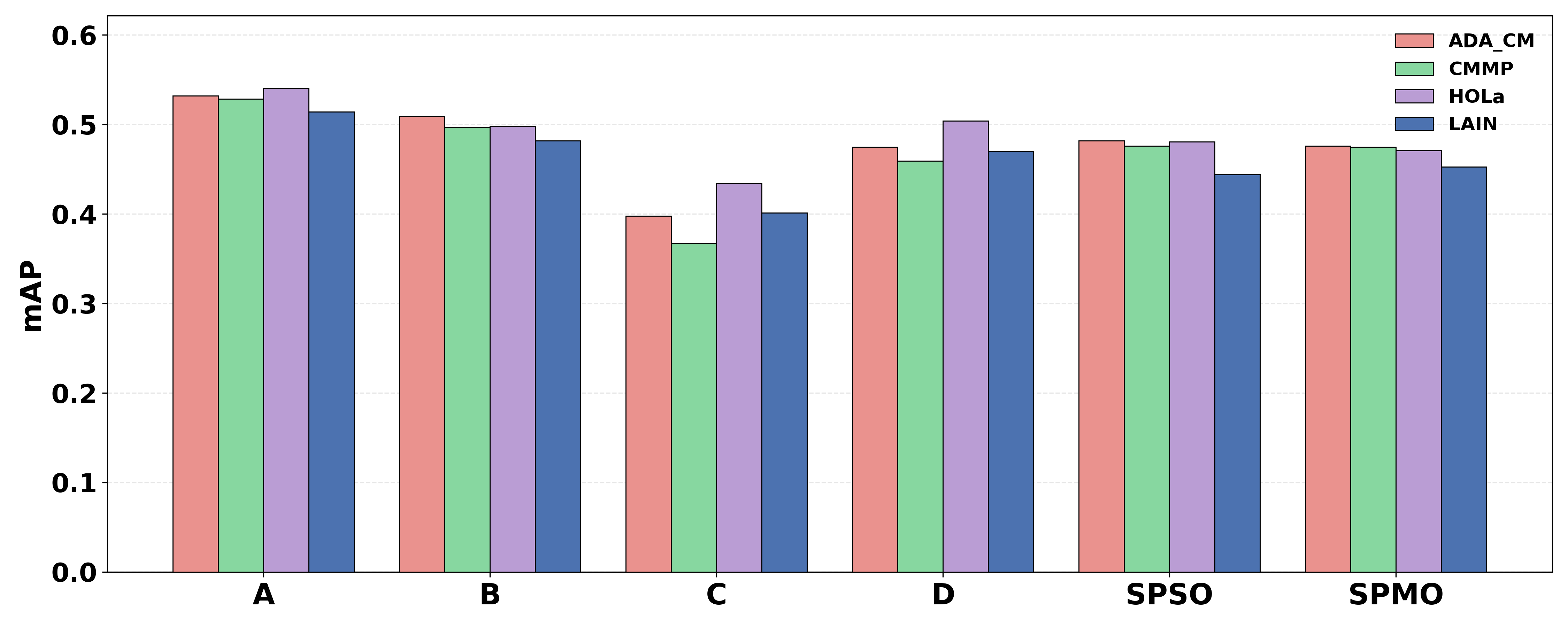}
    \vspace{-2mm}
    \caption{
    mAP across categories.
    While most categories achieve performance comparable to single-person settings (SPSO, SPMO), category C tends to show lower performance across all models, suggesting a consistent source of error.
    }
    \vspace{-3mm}
    \label{fig:category_map}
\end{figure}

\subsection{Performance Gap Between Single- and Multi-Person Scenarios}

\begin{figure*}[t]
    \centering

    \begin{minipage}[b]{0.33\linewidth}
     \subcaption*{ (a) Category A (Multi-Person)} %
        \includegraphics[width=0.99\linewidth]{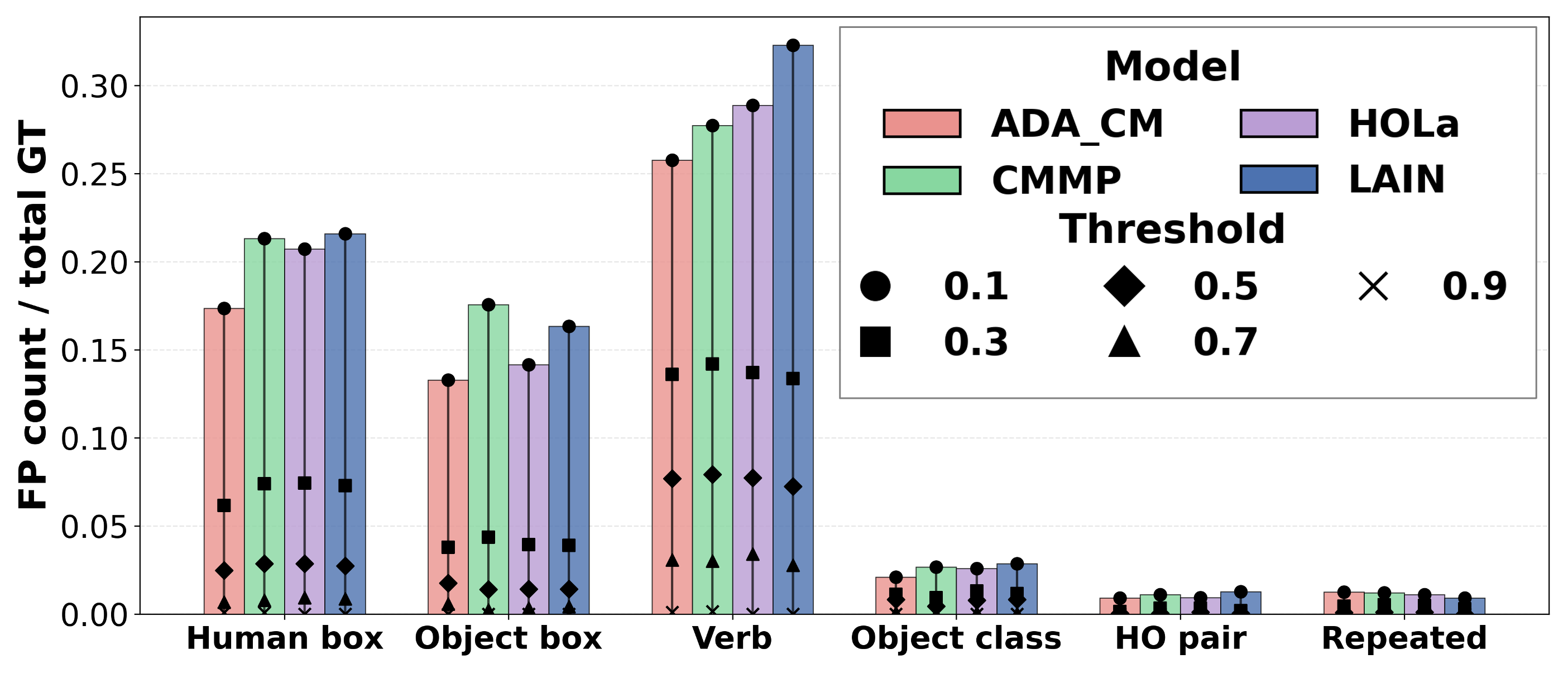}
    \end{minipage}%
    \begin{minipage}[b]{0.33\linewidth}
      \subcaption*{ (b) Category B (Multi-Person)} %
        \includegraphics[width=0.99\linewidth]{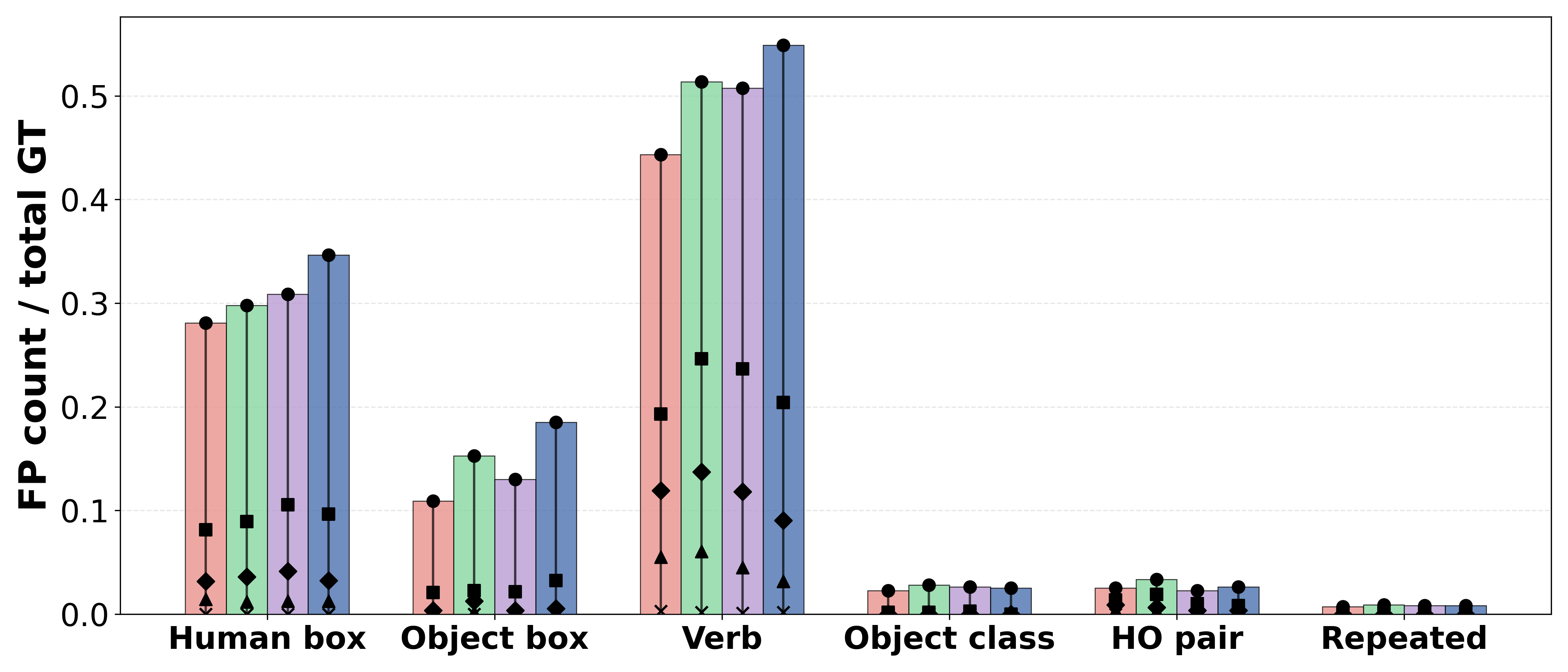}
    \end{minipage}%
    \begin{minipage}[b]{0.33\linewidth}
      \subcaption*{ (c) Category C (Multi-Person)} %
        \includegraphics[width=0.99\linewidth]{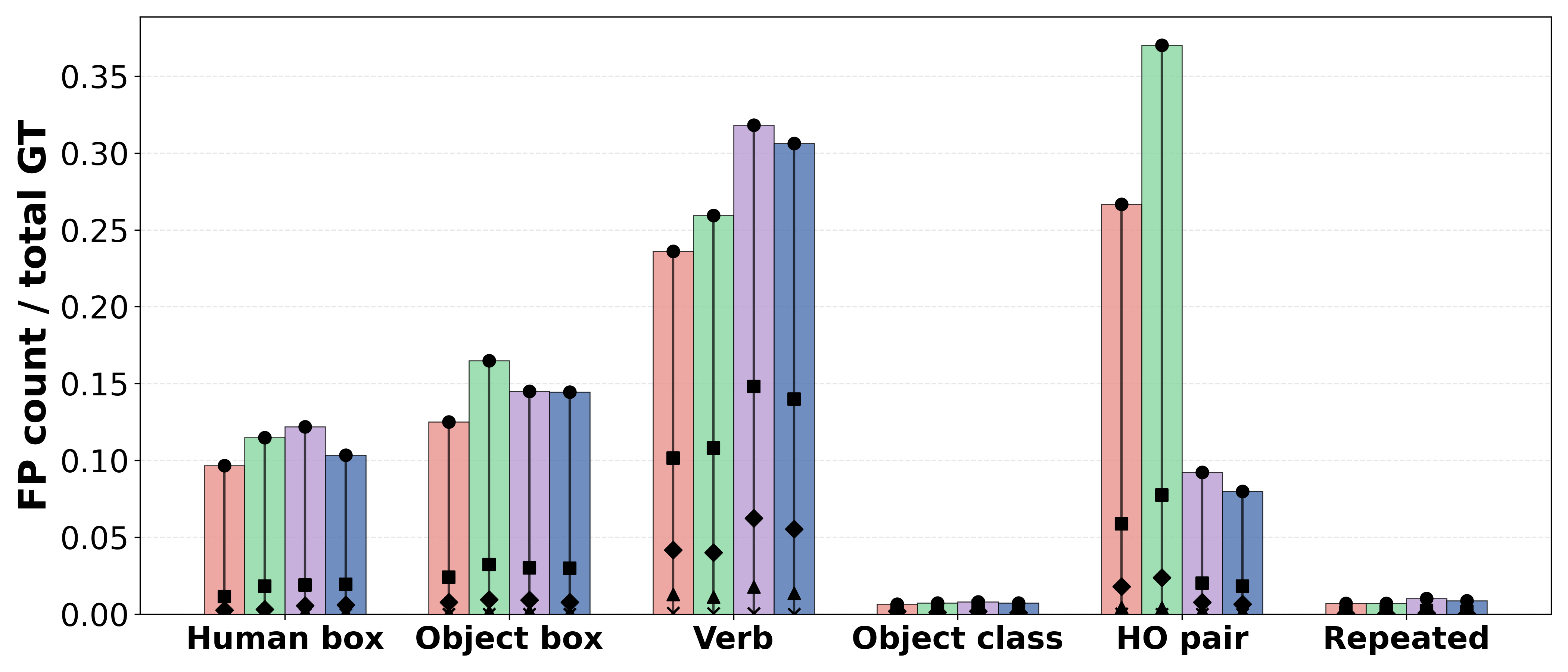}
    \end{minipage}%

    \vspace{1em} 
    \begin{minipage}[b]{0.33\linewidth}
     \subcaption*{ (d) Category D (Multi-Person)} %
        \includegraphics[width=0.99\linewidth]{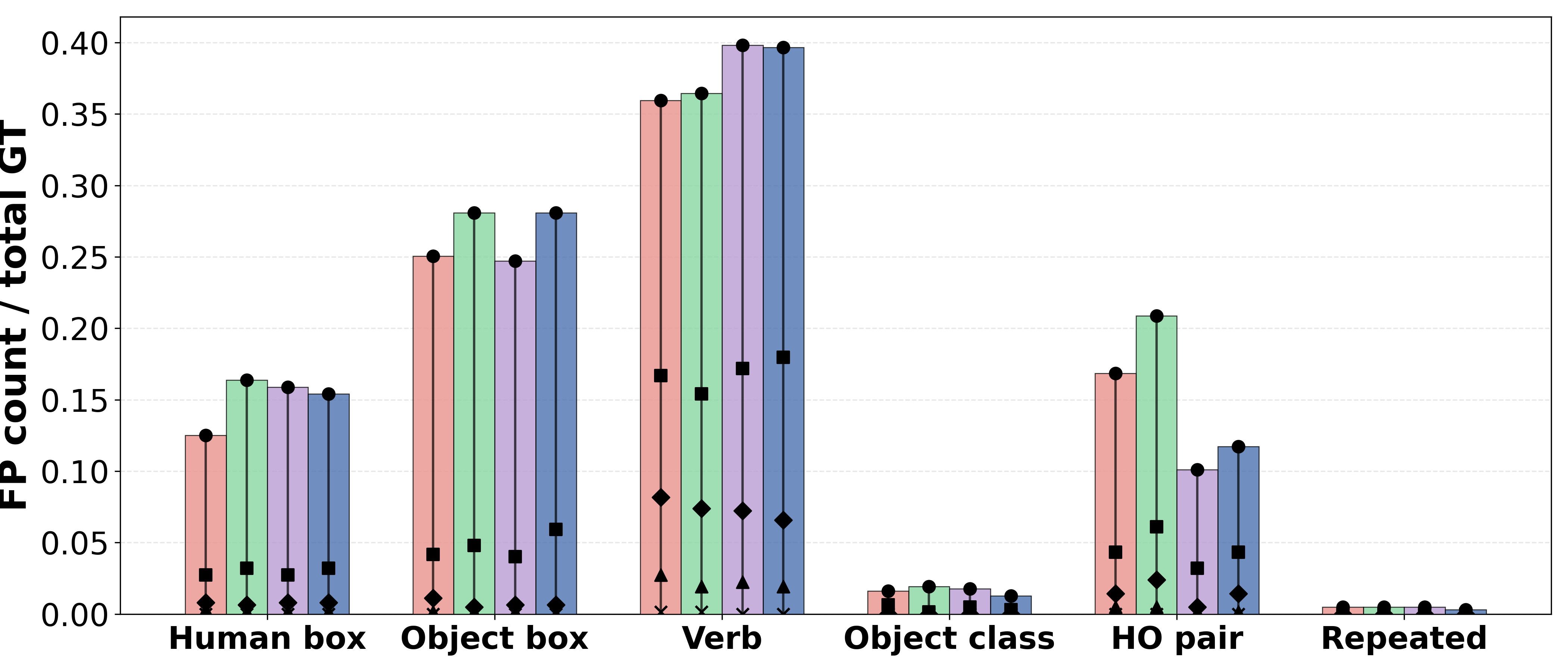}
    \end{minipage}%
    \begin{minipage}[b]{0.33\linewidth}
        \subcaption*{ (e) Category SPSO (Single-Person)} %
        \includegraphics[width=0.99\linewidth]{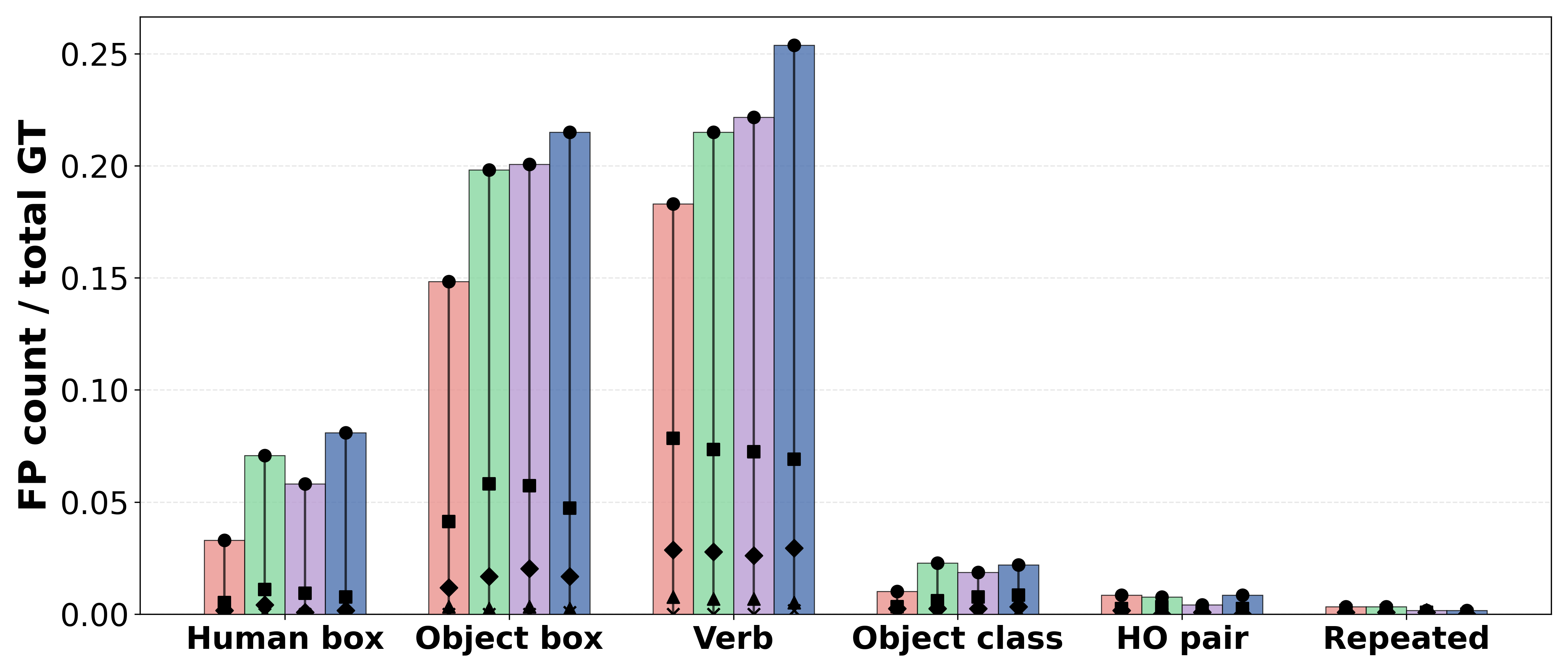}
    \end{minipage}%
    \begin{minipage}[b]{0.33\linewidth}
     \subcaption*{ (f) Category SPMO (Single-Person)} %
        \includegraphics[width=0.99\linewidth]{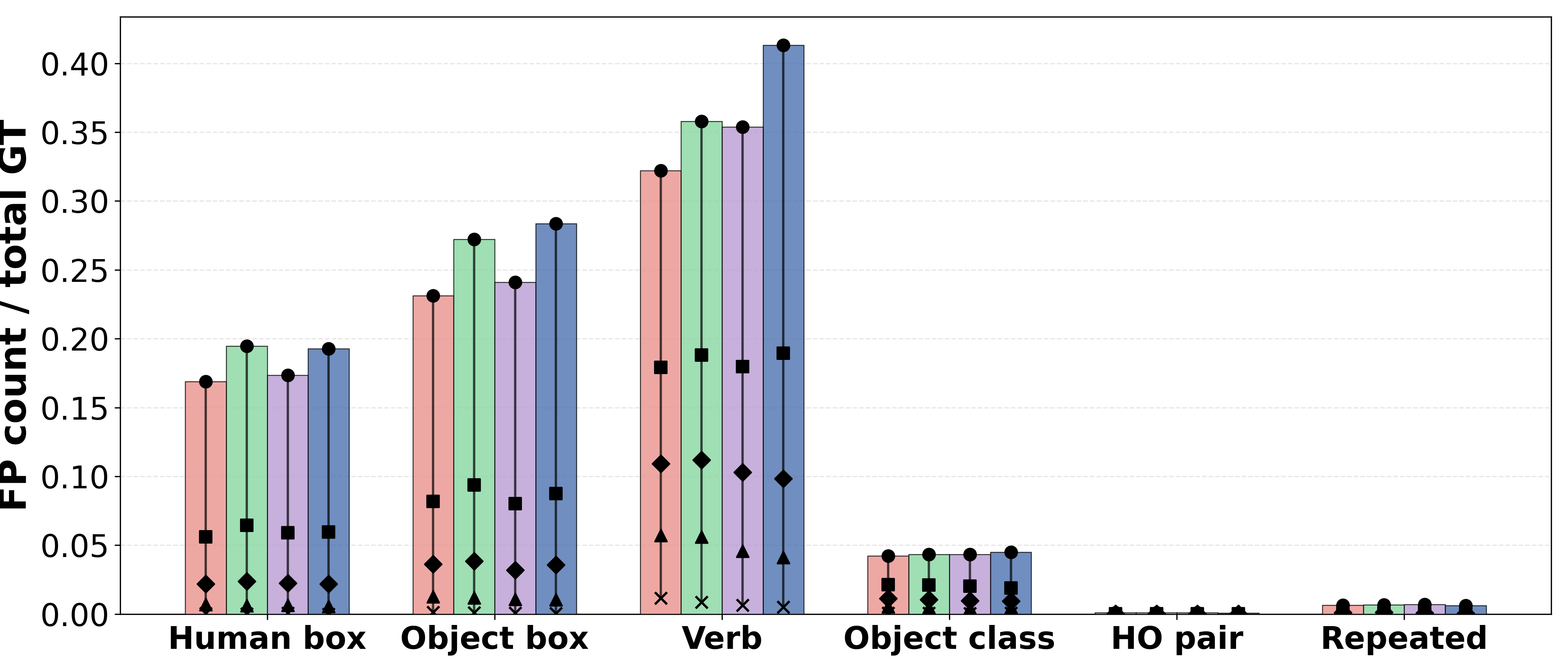}
    \end{minipage}%
    \caption{
    Distribution of false positive error types across categories (A--D, SPSO, SPMO). We decompose incorrect predictions into six types: human box error, object box error, verb classification error, object classification error, human--object pairing error, and duplicate prediction (repeated). These error types are not mutually exclusive, and a single prediction may contain multiple errors.  
    }
    \label{fig:fp_analysis}
\end{figure*}

We first analyze model performance between single-person and multi-person scenarios. As shown in ~\cref{fig:single_vs_multi}, all models show a performance drop when moving from single-person to multi-person settings, suggesting that multi-person scenes are more challenging in this setting.

However, as discussed in ~\cref{Benchmark_stats}, the HICO-DET benchmark is dominated by single-person images (over 60\% of the test set). As a result, standard evaluation is largely influenced by performance on these simpler scenarios, potentially obscuring model behavior in more challenging multi-person interactions. 
Strong overall performance therefore does not necessarily reflect reliable performance in complex multi-person scenarios. This observation suggests the value of a more fine-grained evaluation that separately examines different sources of ambiguity, which we analyze in the following subsections.

\subsection{Performance across Categories}

We further analyze model performance across categories (A--D, SPSO, SPMO), as shown in ~\cref{fig:category_map}. 
We observe that categories A--D do not show substantially lower performance compared to the single-person subsets (SPSO, SPMO). We hypothesize that this is partially due to the reduced number of unique HOIs after restricting the benchmark to images with a single diagnostic category label, as shown in \cref{fig:distribution}. The selection changes the composition of the evaluation set, resulting in interaction patterns that are relatively simpler and less ambiguous.
Moreover, a closer examination shows a pattern: category C exhibits noticeably lower performance across all models. This degradation is consistent across different models, suggesting a consistent source of error rather than random variation.
This may be related to the inherent structure of category C, where multiple objects of the same class are present and different subjects perform the same interaction. 
The presence of multiple objects of the same class introduces instance-level ambiguity, making it difficult for models to correctly associate each human with the corresponding object instance. Further discussions are included in ~\cref{sec:category_analysis}.

\subsection{Error Decomposition Analysis}
\label{sec:category_analysis}
We first define false positive predictions based on standard HOI matching criteria. A prediction is considered correct if both the human and object boxes have IoU greater than 0.5 with a ground-truth pair, and the predicted verb and object classes are correct. Each ground-truth instance is matched to at most one prediction based on confidence ranking, and unmatched predictions are treated as false positives.
Based on this definition, we further decompose false positives into six error types: human box error, object box error, verb classification error, object classification error, human-object pairing error, and duplicate prediction. These error types are not mutually exclusive, and a single prediction may contain multiple errors.

\begin{figure}[t]
    \centering
    \includegraphics[width=0.9\linewidth]{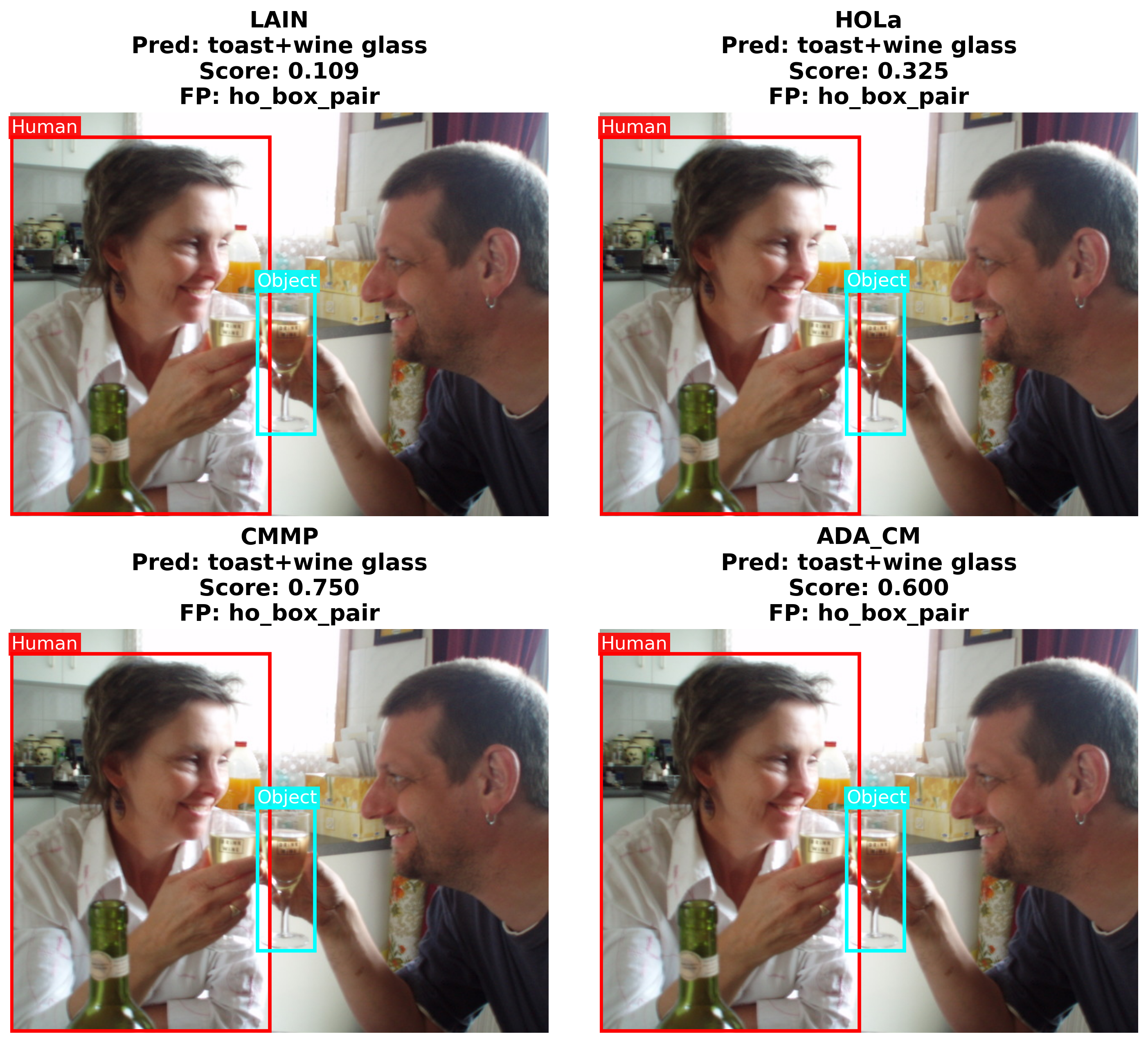}
    \caption{
    Example of human--object pairing errors: the interaction (\textit{toast}) is assigned to the wrong wine glass even when human and object detections are correct across the evaluated models.
    }
    \label{fig:ho_pairing_error_example}
\end{figure}

\noindent \textbf{Human--Object Pairing Errors}
As shown in ~\cref{fig:fp_analysis}, categories C and D show a higher proportion of human--object pairing errors compared to other categories while categories A and B are less affected despite also involving multiple people.
This result indicates that a key challenge in categories C and D lies in instance-level ambiguity, where multiple objects of the same class are present. In such cases, models struggle to correctly associate each human with the corresponding object instance.
This observation is consistent with our organization (\cref{sec:taxonomy}), where categories C and D are designed to isolate ambiguity across object instances with identical semantic labels. 
This suggests that these models may have difficulty resolving fine-grained human--object associations under instance-level ambiguity.
To further illustrate this type of error, we provide an example of human--object pairing errors in ~\cref{fig:ho_pairing_error_example}, where models correctly detect humans and objects but incorrectly associate interactions with the wrong object instance.

We further observe a notable difference across models in the magnitude of pairing errors. 
In particular, models such as HOLa~\cite{lei2025hola} and LAIN~\cite{kim2025locality} exhibit relatively lower pairing error rates compared to others.
One possible explanation lies in how human--object pairs are represented. 
Most evaluated methods follow a conventional two-stage paradigm, where features are extracted by cropping regions from the image. 
In crowded or occluded scenes, such region-based representations may include information from multiple nearby instances, making it more difficult to distinguish between different human--object pairs.
In contrast, HOLa and LAIN construct pair representations using instance-level features from the object detector, which are then used to interact with global image features. 
Such representations may better preserve instance-specific information and be less affected by occlusion, which may contribute to improved pairing accuracy.
While this observation is limited to the models evaluated in this work, it suggests that the quality of instance-level representations may play a role in resolving human--object association ambiguity.

\noindent \textbf{Detection Errors}
In categories A and B, human detection errors remain relatively high even at moderate confidence thresholds (e.g., above 0.5). 
This may be related to the shared-object setting, where multiple people are spatially close and often overlap, making it more challenging to localize individual human instances accurately.
In the SPMO category, object-related errors are more prominent. 
As a single person interacts with multiple objects, the model must select the correct object among several candidates, making object localization more challenging.
These observations further suggest that different scene configurations introduce distinct challenges beyond interaction ambiguity alone. 
Notably, \textbf{verb prediction errors} appear as the most frequent error type across all settings, which we analyze in detail in the following subsection.

\subsection{Verb Prediction Errors}
\label{verb_prediction_errors}
We analyze how verb-related errors vary with different confidence thresholds, as shown in \cref{fig:fp_analysis}. 
While lower thresholds introduce more false positives, we focus on the behavior of errors at higher confidence levels.
As the threshold increases, most low-confidence errors are gradually filtered out. 
However, a non-negligible number of false positives persist even at relatively high confidence (e.g., above 0.5) in several categories, including B, D, and SPMO. 
This suggests that these errors are not only due to uncertain predictions, but may also reflect confident yet incorrect decisions made by the model.

A closer examination shows that these categories share a common structural property: they contain multiple human--object pairs with different but semantically related interactions. 
In categories B and D, different people interact with either the same object instance or different instances of the same object class while performing distinct actions. 
In SPMO, a single person interacts with multiple objects, often with different actions. 
As a result, multiple plausible interaction hypotheses coexist within the same image, leading to increased ambiguity across HO pairs.
Such structured ambiguity can make HOI detection more challenging, as the model must not only localize humans and objects, but also correctly assign the interaction to each pair. 
The persistence of high-confidence errors in these categories suggests that these models may have difficulty distinguishing between closely related interaction patterns, rather than failing due to low confidence.
Overall, these results indicate that threshold adjustment alone may not be sufficient to address errors caused by interaction ambiguity. 
These observations suggest that modeling fine-grained relationships among multiple human--object pairs within a scene may help address such errors.

\begin{figure*}[t]
    \centering

    \begin{minipage}[b]{0.33\linewidth}
     \subcaption*{ (a) Category A (Multi-Person)} %
        \includegraphics[width=0.99\linewidth]{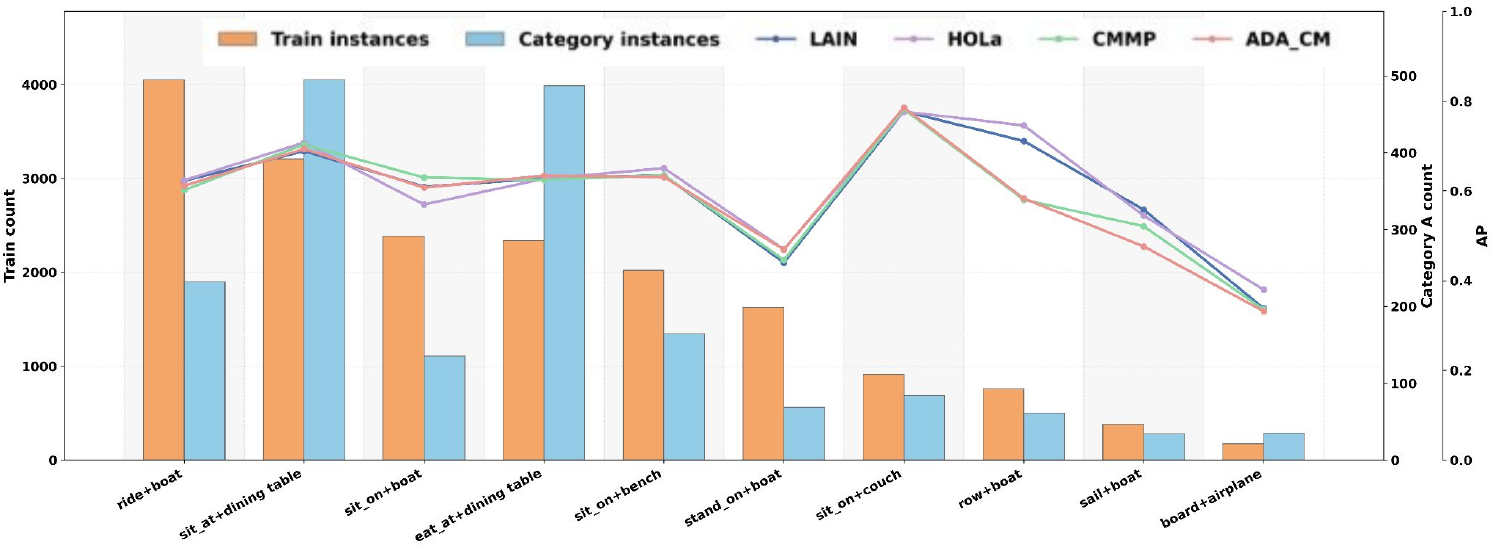}
    \end{minipage}%
    \begin{minipage}[b]{0.33\linewidth}
      \subcaption*{ (b) Category B (Multi-Person)} %
        \includegraphics[width=0.99\linewidth]{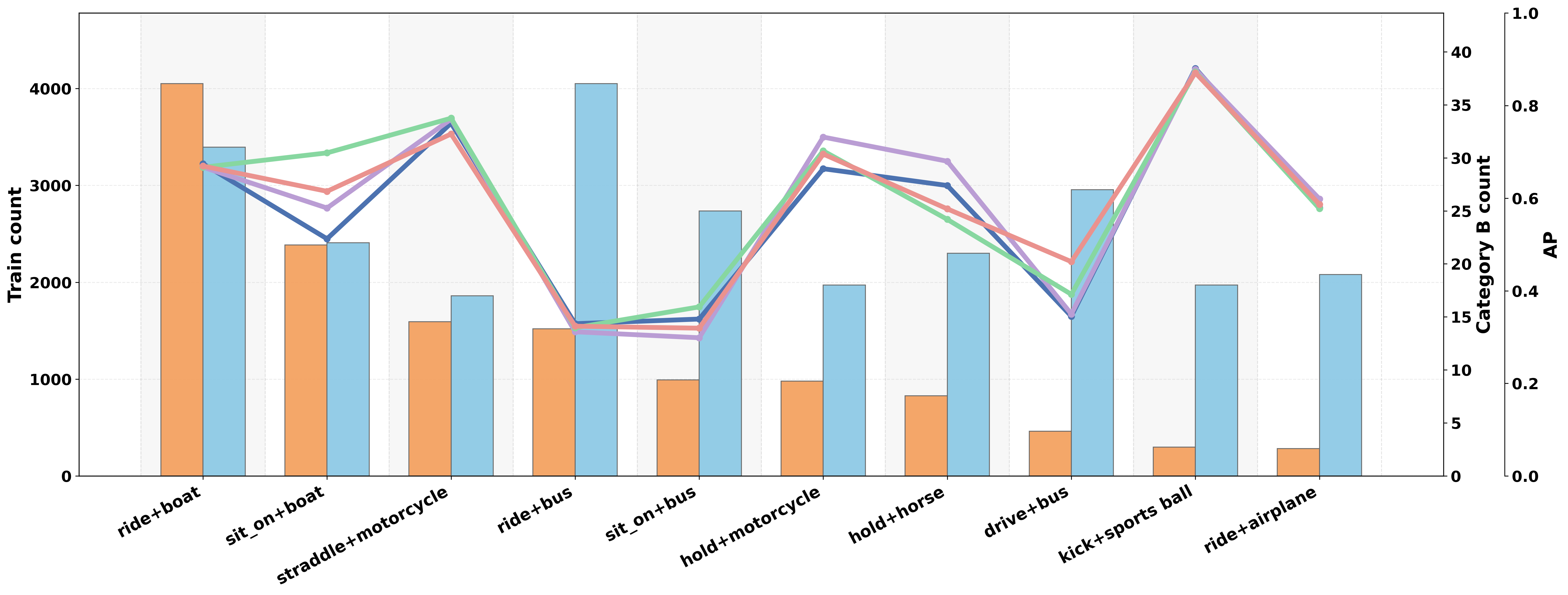}
    \end{minipage}%
    \begin{minipage}[b]{0.33\linewidth}
      \subcaption*{ (c) Category C (Multi-Person)} %
        \includegraphics[width=0.99\linewidth]{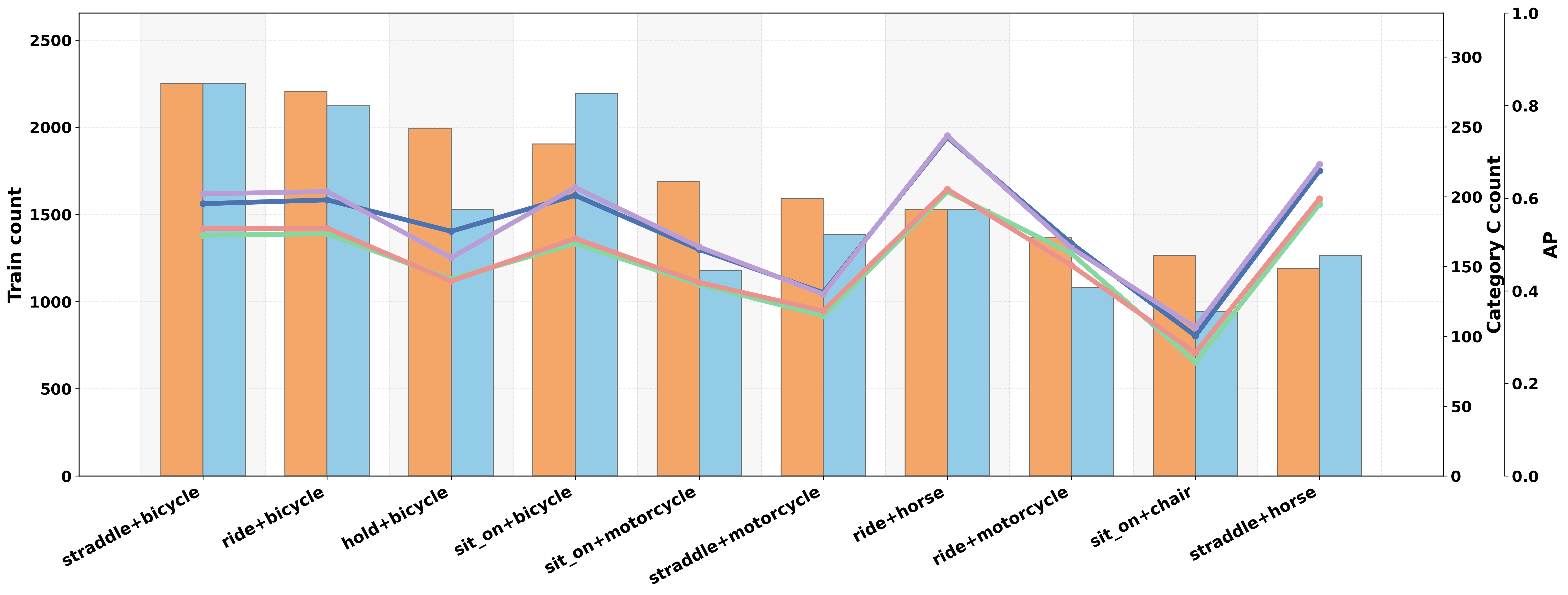}
    \end{minipage}%

    \vspace{1em} 
    \begin{minipage}[b]{0.33\linewidth}
     \subcaption*{ (d) Category D (Multi-Person)} %
        \includegraphics[width=0.99\linewidth]{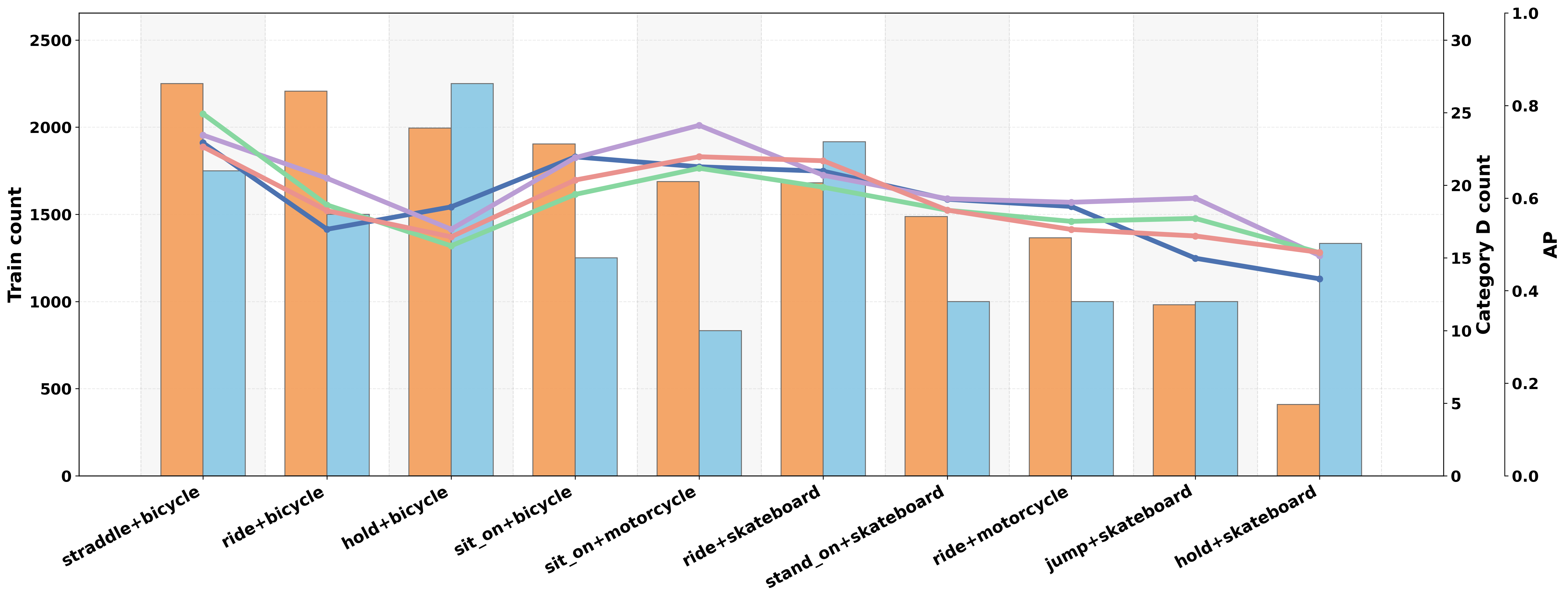}
    \end{minipage}%
    \begin{minipage}[b]{0.33\linewidth}
        \subcaption*{ (e) Category SPSO (Single-Person)} %
        \includegraphics[width=0.99\linewidth]{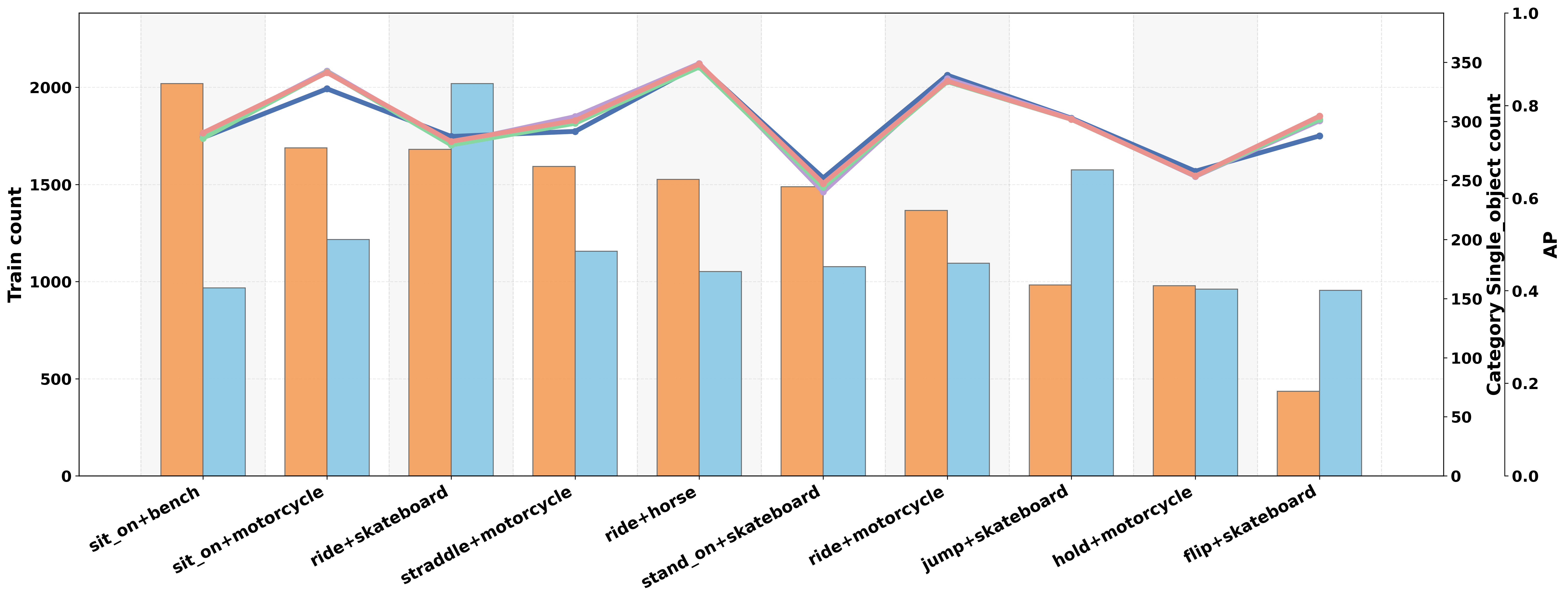}
    \end{minipage}%
    \begin{minipage}[b]{0.33\linewidth}
     \subcaption*{ (f) Category SPMO (Single-Person)} %
        \includegraphics[width=0.99\linewidth]{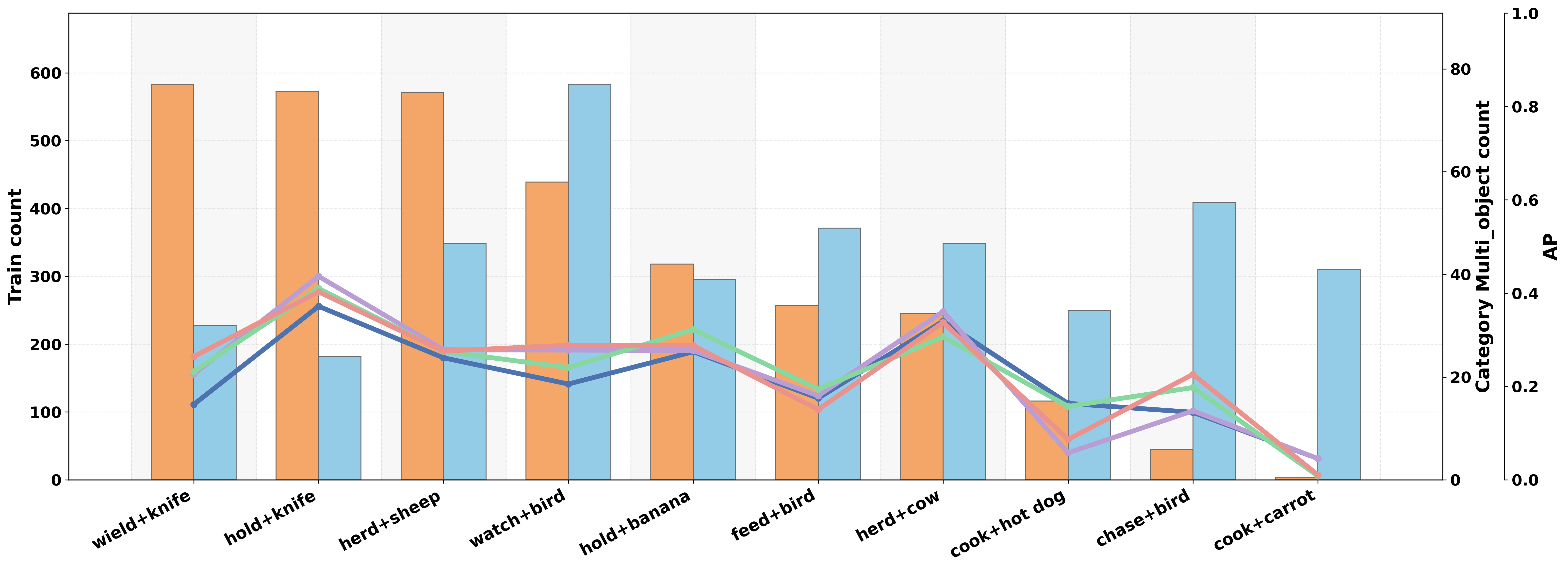}
    \end{minipage}%
     \caption{Top-$10$ HOI distribution and performance across categories. We visualize the top-$10$ HOIs ranked by frequency. Bars indicate the number of training instances (orange) and category-specific occurrences (blue), while lines denote AP across different models.}
    \label{fig:topk_analysis}
\end{figure*}

\begin{figure*}[t]
    \centering
    \begin{subfigure}[t]{0.24\textwidth}
        \centering
        \includegraphics[width=\linewidth]{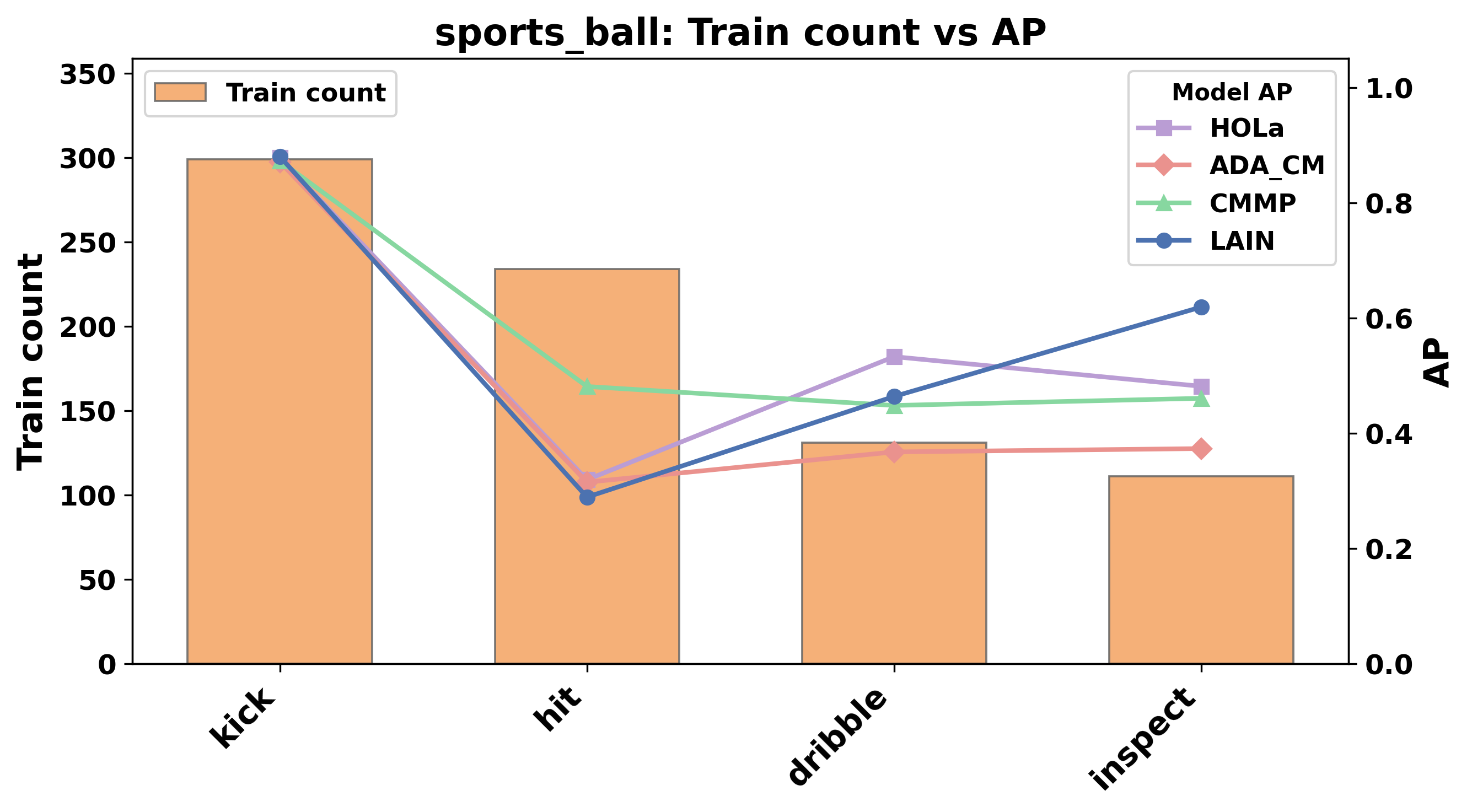}
        \caption{Category B: Sports ball}
        \label{fig:sportsball_pvo}
    \end{subfigure}
    \hfill
    \begin{subfigure}[t]{0.24\textwidth}
        \centering
        \includegraphics[width=\linewidth]{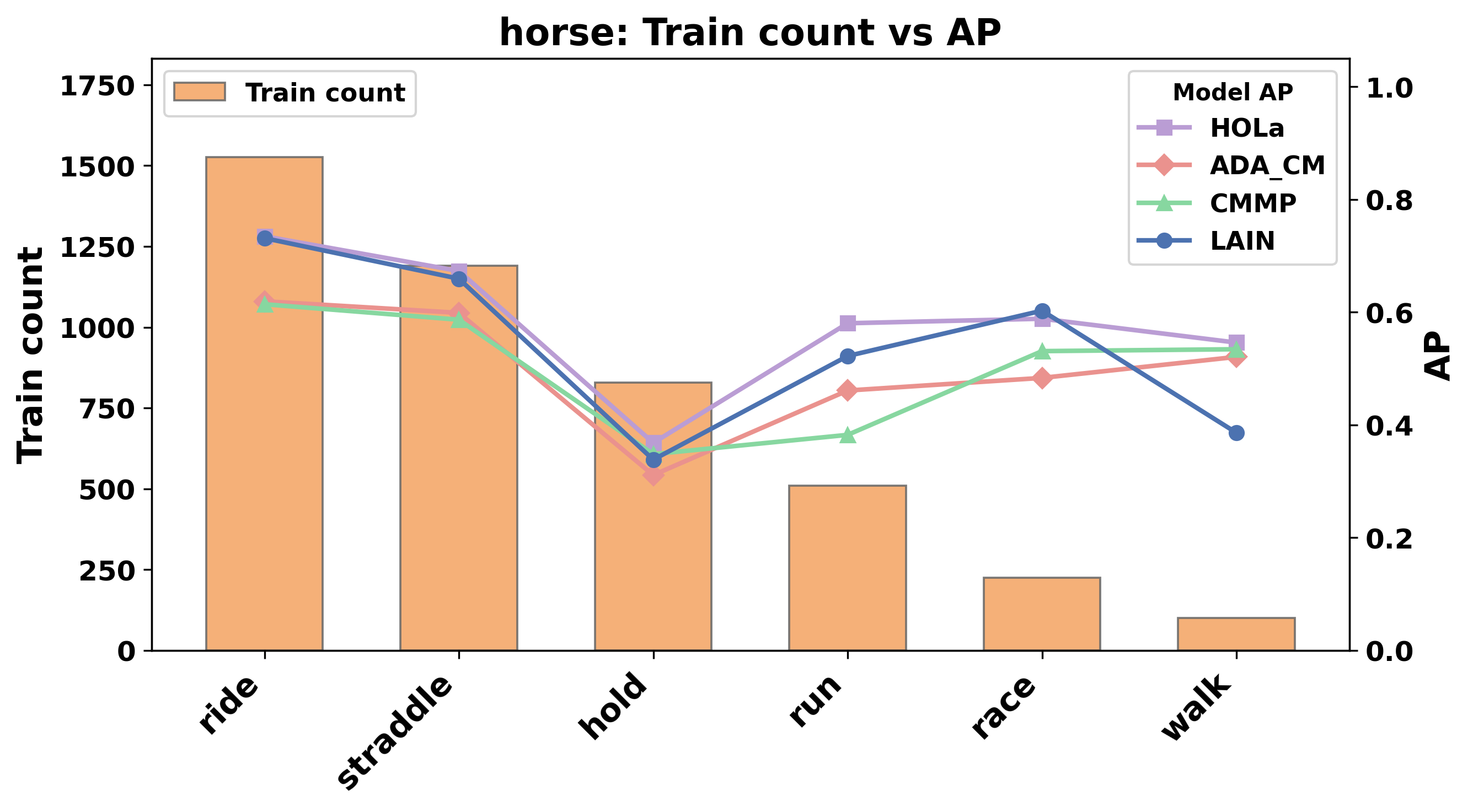}
        \caption{Category C: Horse}
        \label{fig:horse_pvo}
    \end{subfigure}
    \hfill
    \begin{subfigure}[t]{0.24\textwidth}
        \centering
        \includegraphics[width=\linewidth]{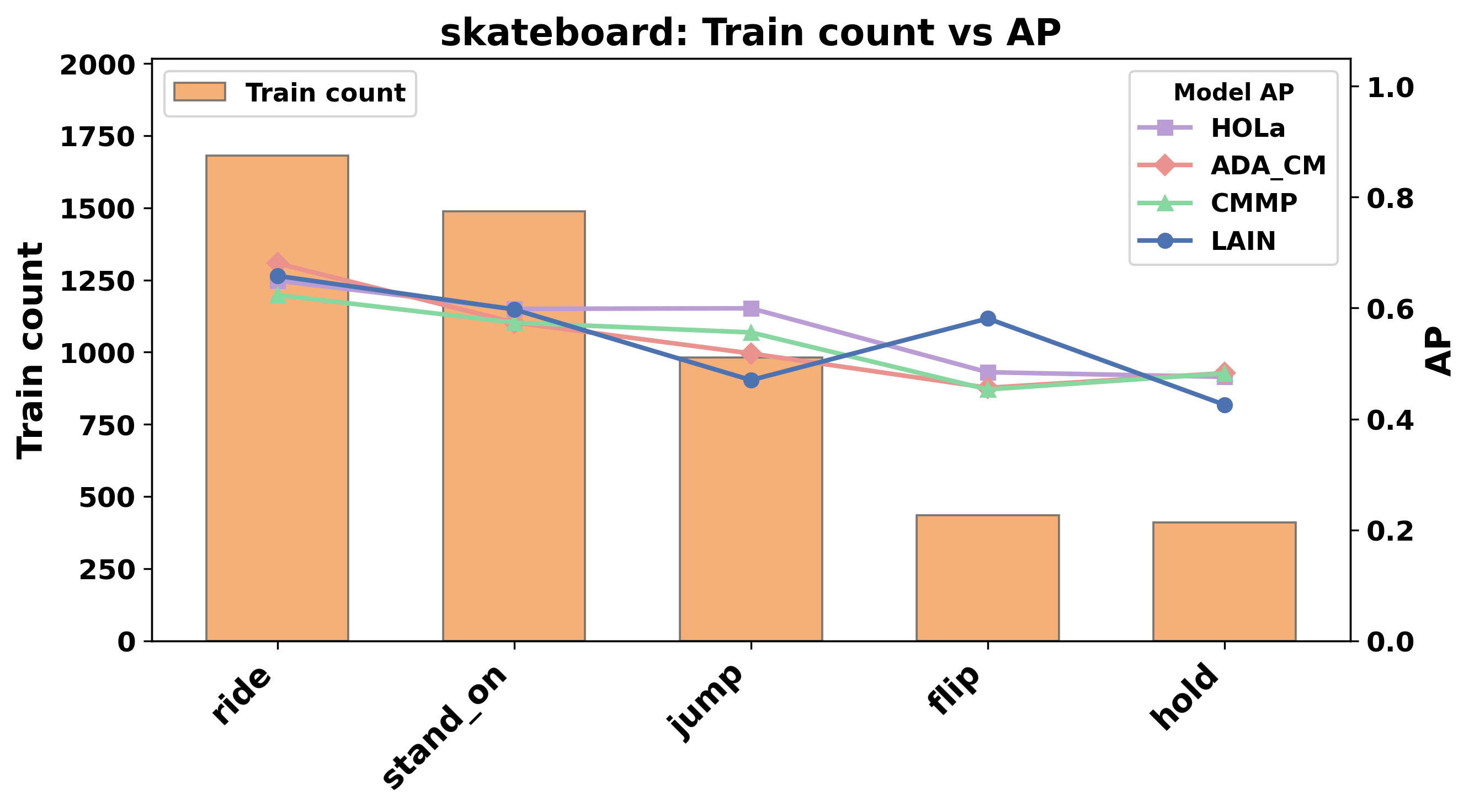}
        \caption{Category D: Skateboard}
        \label{fig:skateboard_pvo}
    \end{subfigure}
    \hfill
    \begin{subfigure}[t]{0.24\textwidth}
        \centering
        \includegraphics[width=\linewidth]{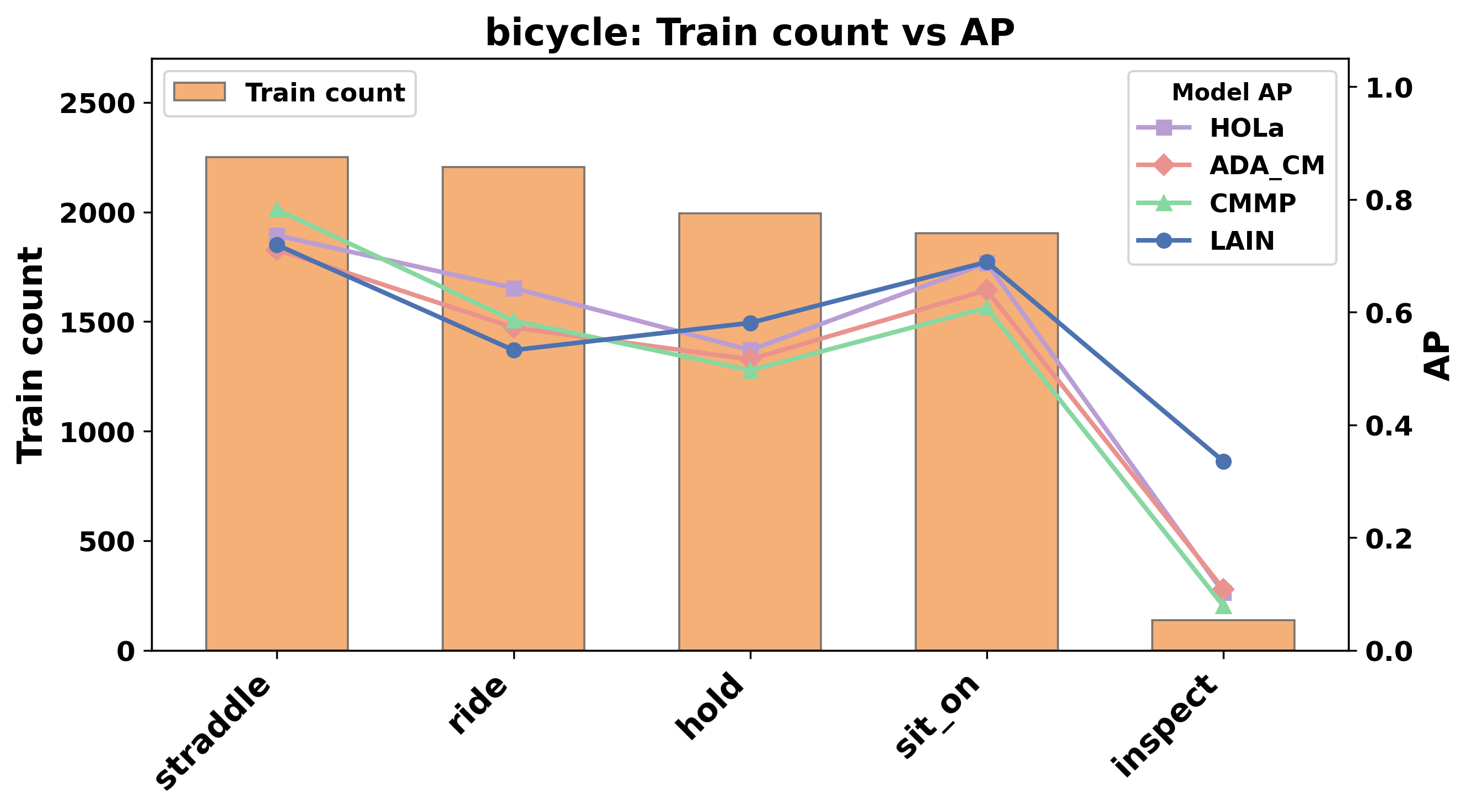}
        \caption{Category D: Bicycle}
        \label{fig:bicycle_pvo}
    \end{subfigure}

    \caption{
    Training verb distributions conditioned on object (orange), measured by instance counts, alongside AP from four models, suggesting object-centric bias in HOI detection.
    }
    \label{fig:pvo_analysis}
\end{figure*}

\subsection{Class-Level Analysis: Object-Centric Bias}
We first analyze the top-10 HOIs in each category to understand how performance varies. 
In \cref{fig:topk_analysis}, we observe the long-tail trend across categories: HOIs with more training instances generally achieve higher AP across models. However, this trend is not absolute. Some HOIs remain strong outliers, achieving high performance despite relatively low frequency in the training set. For example, in category B, \textit{kick sports ball} yields high AP across all models even though its instance count is limited. 
This suggests that HOI performance may not be determined solely by the frequency of each HOI class itself.
Instead, we hypothesize that the performance may also depend on object-conditioned bias, where predictions are biased toward verbs that dominate among interactions of a given object in the training data. In other words, even if a HOI class such as \textit{kick sports ball} is not frequent in the overall training set, it may still achieve high performance if it constitutes a large proportion of interactions involving that object (e.g., among \textit{sports ball} instances). As a result, once the object is detected, the models may favor verbs that are dominant for that object.

To further investigate this phenomenon, we analyze verb distributions conditioned on object classes and relate them to model performance, as shown in \cref{fig:pvo_analysis}. 
For each object, we visualize the number of training instances for each verb (bars) together with the corresponding per-HOI AP across models (lines). 
We focus on four representative objects, \textit{horse}, \textit{sports ball}, \textit{skateboard}, and \textit{bicycle}. We only consider HOI classes that have at least 5 test instances in the corresponding categories to ensure statistically reliable analysis.
We observe a consistent trend that, within each object, verbs with higher training frequency tend to achieve higher AP, while less frequent verbs perform worse. 
More importantly, this correlation appears at the object level rather than at the global HOI level, suggesting that model predictions may be influenced by object-conditioned verb distributions. 
In other words, models may favor verbs that dominate among interactions of a given object, suggesting reliance on object–verb co-occurrence patterns.

\subsection{Limitations}
This study has several limitations. 
First, our analysis is based on HICO-DET annotations, which do not provide instance-level identity, limiting fine-grained analysis of human–object associations in complex scenes. 
Second, we restrict our analysis to images with a single category label, excluding cases with overlapping configurations. 
Addressing these limitations would enable analysis under more complex and realistic interaction scenarios.

\section{Conclusion}
\label{sec:conclusion}
In this work, we present a study to analyze the failure modes of two-stage HOI detection methods.
We analyze HOI performance across human--object interaction configurations and error types, complementing standard metrics such as mAP.
Our analysis suggests that different configurations induce distinct failure patterns, with verb prediction emerging as a dominant source of error and many errors persisting even at high confidence.
We hope that our study provides useful information for developing more robust HOI detection models in future work.

\clearpage
{\small
\section*{Acknowledgments}
We are grateful for the support of the Ohio Supercomputer Center for providing computational resources. 
This work was supported in part by the Mississippi Impact Grant (MIG), Office for Research and Economic Development, University of Mississippi.
}
{
    \small
    \bibliographystyle{ieeenat_fullname}
    \bibliography{main}
}


\end{document}